\RequirePackage{fix-cm}

\documentclass{article} 
\usepackage{arxiv}
\usepackage{natbib}
\usepackage{amsfonts}
\usepackage{amstext}
\usepackage{amsmath}
\usepackage{rotating}

  \usepackage{graphicx}
  \usepackage{mathptmx}      % use Times fonts if available on your TeX system
  \newcommand{\R}{\mathbb{R}}

	\title{Methods for Pruning Deep Neural Networks}
	
	\author{Sunil Vadera \\
		University of Salford,\\
		Greater Manchester, M5 4WT, UK \\
		\texttt{S.Vadera@salford.ac.uk} \\
		\And
		 {Salem Ameen}\\
		 University of Salford,\\
		 Greater Manchester, M5 4WT, UK \\
		\texttt{S.Ameen@edu.salford.ac.uk}} 
	
\begin{document}

\maketitle
 
\begin{abstract}

This paper presents a survey of methods for pruning deep neural networks. It begins by categorising over 150 studies based on the underlying approach used and then focuses on three categories: methods that use magnitude based pruning, methods that utilise clustering to identify redundancy, and methods that use sensitivity analysis to assess the effect of pruning. Some of the key influencing studies within these categories are presented to highlight the underlying approaches and results achieved.
Most studies present results which are distributed in the literature as new architectures, algorithms and data sets have developed with time, making comparison across different studied  difficult. The paper therefore provides a resource for the community that  can be used to quickly compare the results from many different methods on a variety of data sets, and a range of architectures, including AlexNet, ResNet, DenseNet and VGG. The resource is illustrated by comparing the results published for pruning AlexNet and ResNet50 on ImageNet and ResNet56 and VGG16 on the CIFAR10 data to reveal which pruning methods work well in terms of retaining accuracy whilst achieving good compression rates. The paper concludes by identifying some promising directions for future research.

\end{abstract}

\maketitle

\section{INTRODUCTION}
\label{section:Introduction}
Deep learning and its use in high profile applications such as autonomous vehicles~\citep{kuutti2019}, predicting breast cancer~\citep{McKinney2020}, speech recognition~\citep{Hinton2012a} and natural language processing~\citep{otter2021}  have propelled interest in Artificial Intelligence to new heights, with most countries making it central to their industrial and commercial strategies for innovation.

Although there are different types of architectures~\citep{Pouyanfar2019}, deep networks typically consist of layers of neurons that are connected to neurons in preceding layers via weighted links. Another characteristic, which is considered central to their predictive power~\citep{Sejnowski2020},  is that they have a large number of parameters that need to be learned, with networks such as ResNet50 \citep{He2016} having more than 25 million parameters and VGG16~\citep{Simonyan2015}  having more than 138 million weights.   
An obvious question, therefore, is to ask whether it is possible to develop smaller, more efficient networks without compromising accuracy?  One direction of work aimed at addressing this question has been to first train a large network and then to prune and fine-tune a network. Although methods for pruning shallow neural networks were proposed in the 1980s and 90s~\citep{Mozer1988, Kruschke1988,Reed1993}, recent advances in deep learning and its potential for applications in embedded systems has led to an increasing number and variety of algorithms for pruning deep neural networks. Hence, this paper presents a survey of recent work on pruning neural networks that can be used to understand the types of algorithms developed, appreciate the key ideas underpinning the algorithms and gain familiarity with the major approaches and issues in the field.  The paper aims to achieve this goal by presenting the progressive path from the earlier algorithms to the recent work, categorising algorithms based on the approach used, highlighting the similarities and differences between the algorithms and concluding with some directions for future research.  

The studies on pruning methods all carry out empirical evaluations that compare the performance of algorithms on different architectures and benchmark data sets.  These  evaluations have evolved as new deep learning architectures have developed, as new data sets have become available and as new pruning algorithms have been proposed. This paper also provides and illustrates the use of a resource that brings together the reported results in one place, allowing researchers to quickly compare the reported results on different architectures and data sets.

The survey identified over 150 studies on pruning neural networks, which can be categorised into the following eight groups based on the underlying approach used:
\begin{enumerate}
\item {\bf Magnitude based pruning methods} \citep{ Chauvin1988,Weigend1990,Weigend1991,Zhou2018}, which are based on the view that  the saliency of weights and neurons can be determined by local measures such as their magnitude or approximated by their effect on the next layer. 
\item {\bf Similarity and clustering methods}~\citep{Chen1993, Han2016, Li2019b,RoyChowdhury2017,Sussmann1992,Zhou2018a} which aim to identify duplicate or similar weights which are redundant and can be pruned without impacting accuracy.
\item {\bf Sensitivity analysis methods}~\citep{Mozer1988,LeCun1990,Hassibi1993,Hassibi1993a,Cohen2016, Lee2018,Lin2018}, that assess the effect of removing or perturbing weights on the loss and then remove a proportion of the weights that have least impact on accuracy.
\item {\bf Knowledge distillation methods}~\citep{Bucilua2006,Hinton2015,GregorUrban2017,Zhang2019a} which utilise the original model, termed the Teacher, to  learn a more compact new model called the Student.
\item {\bf Low rank methods}~\citep{Sainath2013,Jaderberg2014,Lin2018}  that factor a weight matrix  into a product of two smaller matrices which can then be used to perform an equivalent function more efficiently  than the single larger weight matrix.
\item {\bf Quantization methods}~\citep{Jung2019,Zhou2017a,Zhao2019,Courbariaux2016,Chen2015,Jacob2018},  which are based on using quantization, hashing, low precision and binary representations of the weights as a way of reducing the computations.
\item {\bf Architectural design methods}~\citep{Baker2017, Dai2019, Li2019a,Liu2019metapruning, Lin2020a, Zhong2018, Zoph2017} that utilise intelligent search and reinforcement learning methods to generate neural network architectures.
\item {\bf Hybrid methods}~\citep{Chung2016,Goetschalckx2018,Gadosey2019}  which utilise a combination of methods aimed at taking advantage of the cumulative compressing effects of the different types of methods.
\end{enumerate}

\begin{figure}
\centering
\includegraphics[width=6in]{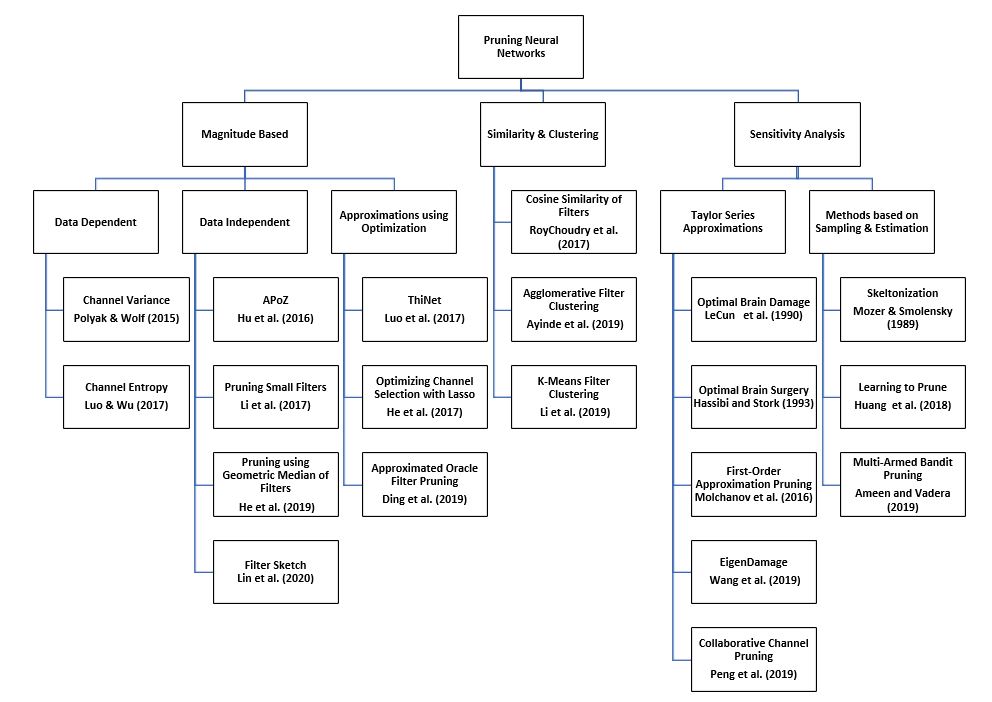}
\caption{A selection of pruning methods grouped in terms of the approach adopted}
\label{figure:categorisation}  
\end{figure}
Appendix A provides a table which classifies the existing studies into the 8 categories, enabling researchers working on a particular type of method to locate related studies. Given the range of studies, and availability of surveys already covering some of the above categories, this paper focuses on recent algorithms in the first three categories for pruning.  \citet{Reed1993} provides an excellent survey of pruning methods prior to the deep learning era. Readers interested in the use of quantization, low rank and knowledge distillation methods are referred to the  survey by \citet{Lebedev2018} and readers interested in architectural design  methods are referred to the comprehensive survey by \citet{Elsken2019}.  Pruning networks is just one step in developing efficient models and a recent survey by~\citet{Menghani2021} summarises the full range of methods, from use of quantization and learning, to the available  software and hardware infrastructure for efficient deployment of models. 
Another important direction of work, worthy of a survey in its own right, and not in the scope of this paper, is the use of variational Bayesian methods for regularization ~\citep{Arbib2003,Goodfellow2016,Huang2018,Lin2020b,Srivastava2014,Wen2016,Zhao2019a}.

Fig.~\ref{figure:categorisation} shows a selection of the methods covered in greater detail in this survey and includes a  sub-categorization of magnitude and sensitivity analysis methods.  The survey found relatively few methods that utilise similarity and clustering, and further sub-categorization is not useful.  
Magnitude based methods can be sub-categorised into: (i) data dependent methods that  utilise a sample of examples to assess the extent to which removing weights impacts the outputs from the next layer; (ii) data independent methods, that  utilise measures such as the magnitude of a weight; and (iii) the use of optimisation methods to reduce the number of weights in a layer whilst approximating the function of the layer.  Methods that utilise sensitivity analysis can be sub-categorized into those that:  (i) adopt a Taylor series approximation of the loss and  (ii) use sampling to estimate the change in loss when weights are removed.  

The rest of this paper is organised as follows.  Section~\ref{section:background} presents the background. Sections~\ref{section:magnitude_pruning}
to \ref{section:sensitivity} describe representative methods in the three categories: magnitude based pruning, clustering and similarity, and sensitivity analysis. Section~\ref{section:magnitude_pruning} also includes coverage of the Lottery Hypothesis, an issue about the existence of smaller networks and fine-tuning, that cuts across the different methods. Section~\ref{section:resource}  presents a comparison of the published results for pruning AlexNet, ResNet and VGG to illustrate the resource provided for comparing the methods. Section~ \ref{section:conclusion} concludes by highlighting some key insights and suggesting directions for future research.

\begin{figure}

\centering
\includegraphics[width=4.5in]{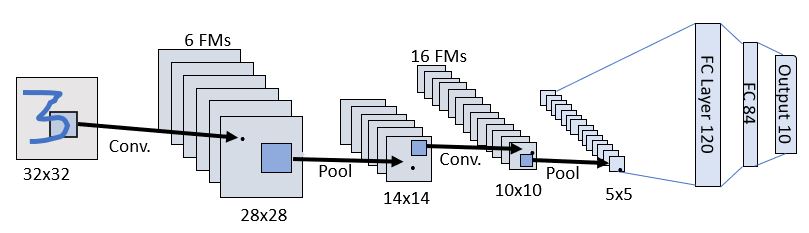}

\caption{The LeNet-5 network and how it processes an input image via convolutions (Conv.) and pooling operations to produce features maps (FMs) and uses fully connected (FC)  layers  to perform classification}
\label{figure:lenet}

\end{figure}

\section{BACKGROUND}

\label{section:background}
 
This section introduces the background knowledge assumed in the survey.\footnote{Readers unfamiliar with deep neural networks are referred to tutorial accounts such as~\citep{Goodfellow2016} for further details}  
 Fig.~\ref{figure:lenet} shows the structure of one of the earliest convolutional neural networks (CNNs), LeNet-5~\citep{LeCun1989a}, which recognises handwritten digits by  applying convolutions and pooling operations to identify features.  These features then provide the input to fully connected layers that classify the images. The pooling operation takes feature maps as input and reduces their size by applying an operation, such as the maximum value within a neighbourhood while the convolution operation applies filters (or kernels) to the input channels (or feature maps) to produce the output feature maps. The filters are k x k matrices that slide over the input feature maps and convolve with the corresponding elements of the input feature maps to produce the output feature maps.  The elements of a filter correspond to the weights (or parameters) that are used to transform regions in feature maps in one layer to the next and need to be learned through training. The weights (or parameters), either individually or collectively as filters, are therefore the primary candidates for pruning.   

The LeNet-5 model, with 60K parameters in 5 layers, achieved impressive results on a data set known as MNIST ~\citep{LeCun1998a}.\footnote{Modified National Institute of Standards and Technology}   In a breakthrough in 2012, AlexNet built upon the concepts in LeNet-5  and developed a deeper network with over 60M parameters in  8 layers  to win a competition known as ImageNet by a significant margin~\citep{Krizhevsky2012}. This success was followed by the development of architectures like VGG, ResNet, and ResNeXT that used an increasing number of layers and parameters to gain further improvements in the ImageNet competition~\citep{Pouyanfar2019}. 
The huge number of parameters in these models does necessitate greater computational resources and inhibits their use in embedded systems, which has motivated the research on pruning that is surveyed in this paper.  

The pruning methods developed are evaluated on a range of architectures (e.g., ResNet, VGG, DenseNet) and data sets (e.g., ImageNet CIFAR, SVHN).
\citet{Khan2020} presents a tutorial on deep learning architectures and Appendix B summarises the data sets. When evaluating pruning methods, the surveyed papers use the following measures to report their results:
\begin{itemize}
	\item The Top-1 and Top-5 accuracy, which report the proportion of times the correct classification appears first or in the top 5 list of ranked results. In the sections below, unless we explicitly qualify a measure, the Top-1 accuracy should be assumed.
	\item The compression rate, which is the ratio of parameters before and after a model is pruned.
	\item The computational efficiency in terms of the FLOPS (Floating Point Operations) required to perform a classification.
\end{itemize}
The notation used in the paper is defined where it is used and also summarised in Appendix C.
With this background in place, Sections~\ref{section:magnitude_pruning} to \ref{section:sensitivity}
 describe some key  influential studies  that bring out the features of the categories of methods surveyed in this paper.

\section{MAGNITUDE BASED PRUNING}
\label{section:magnitude_pruning}

This section presents pruning methods that remove weights, nodes, and filters based on a measure of magnitude or  the effect filters have on the next layer.  Subsection~\ref{subsectiona:netpruning} summarizes an early influential method for pruning weights and Subsection~\ref{subsection:lottery} presents a recent hot topic, termed the Lottery Hypothesis,  that reinvigorates research on the existence of smaller networks and raises issues about fine-tuning a pruned network.  Subsection~\ref{subsectionc:featuremaps} describes the key ideas behind methods that prune filters and feature maps.

\subsection{Network Pruning of Weights}

\label{subsectiona:netpruning}
One of the first studies to utilise magnitude based pruning for deep networks is due to \citet{Han2015a} who adopt a process in which weights below a given threshold are pruned.  Once pruned, the network is fine-tuned and the process repeated until its accuracy begins to deteriorate.   

\citet{Han2015a} carry out several experiments to compare the merits of their magnitude based iterative pruning method. First, they apply their method on a  fully connected network known as LeNet-300-100  and then on Lenet-5 (Fig.~ \ref{figure:lenet}), both of which are trained on the MNIST data.  Their results show that it is possible to reduce the number of weights by a factor of 12 without compromising accuracy. Second, they apply iterative pruning to AlexNet and VGG16 trained on the ImageNet data, and show that it is possible to reduce the number of weights by a factor of 9 and 12 respectively.  Thirdly, they compare the merits of using regularisation to drive down the magnitude of weights to aid subsequent pruning.  They explore regularisation with both the $L_1$ and $L_2$ norms and conclude that $L_1$ is better immediately after pruning (without fine-tuning), but $L_2$ is better if the weights of the pruned model are fine-tuned.  Their experiments also suggest that the earlier layers (i.e., closer to the inputs) are the most sensitive to pruning and that iterative pruning is better than pruning the required proportion of weights in one cycle (i.e., one-shot pruning).  

The study by~\citet{Han2015a} is notable in that (i) it demonstrated that it was possible to  reduce the size of deep networks significantly without compromising accuracy, (ii) it highlighted the benefits of iterative pruning and (iii) it prompted further research on questions such as whether retraining from scratch or fine-tuning is better following pruning.

\citet{Guo2016} note that magnitude pruning can lead to premature removal of weights that can become important given removal of other weights. To address this, they propose a method known as Dynamic Network Surgery (Dyn Surg)  which maintains a mask that indicates which weights should be removed and retained in each training cycle, thereby allowing reinstatement of weights previously marked to be pruned if they turn out to be important.  \citet{Guo2016}  compare their method with magnitude pruning, with the results showing that it is reduces the number of weights by a factor of over 17 for AlexNet on ImageNet.

\subsection{The Lottery Ticket Hypothesis}
\label{subsection:lottery}

One of the most interesting observations in ~\citep{Han2015a} is that re-initialization of the weights does not lead to accurate models and, based on their trials,  it was better to fine-tune the weights of the pruned model.  Following on from this observation, \citet{Frankle2019a} propose the Lottery Ticket Hypothesis which states that: a trained network contains a subnetwork, which can be trained to be at least as accurate as the original network using no more than the number of epochs used for training the original network. This subnetwork is termed a winning lottery ticket, given that it was lucky to be initialised with suitable weights.

To test this hypothesis, they propose two pruning methods. First, in a one-shot method, they use magnitude pruning to prune p\% of the weights, reset the remaining weights to their initial values and retrain.  Second, they utilise an iterative pruning method with $n$ cycles, with each cycle pruning  $p^{1/n}$ of the weights.  

They perform experiments on the fully connected LeNet-300-100 network for the MNIST data, and  variants of VGG and ResNet for the CIFAR10 data. 
Their experiments on the LeNet-300-100 network prune a percent of the weights from each layer except the final layer, in which the percent pruned is reduced by half.  Their results with iterative pruning show that:  (i) a subnetwork that is only 3.6\% of its original size performs just as well,  (ii) random initialization of the pruned networks results in slower learning in comparison to use of the original weight initializations, (iii) that the subnetworks (termed winning tickets) found, learn faster than the original network,  (iv) there is  continual improvement in the rate of learning as the size of the network reduces, but only up to a point, after which learning slows down and begins to regress to the performance of the original network, (v) iterative pruning tends to result in more accurate smaller networks than one-shot pruning.

Their experiments  on the larger networks, VGG and ResNet, show that identification of winning lotteries depends on the learning rate, with a lower rate successfully identifying winning lottery subnets, and that pruning weights over all the network, as opposed to layer by layer produces better results.

These results provide good empirical evidence for the Lottery Hypothesis and the award of a best paper prize in the 2019 International Conference on Learning Representations is indicative of the significance of the paper and the attention it has attracted. 

In their paper, ``Rethinking the value of network pruning'', \citet{Liu2019} challenge the claim that it is better to utilise the initial weights of a pruned model when compared with random initialization. To test this, they carry out experiments on VGG, ResNet, and DenseNet using the CIFAR10, CIFAR100, and ImageNet data. They define three types of pruning regime:  structured pruning, where the proportion of channels that are pruned per layer is predefined; automatic pruning, where the proportion of channels pruned overall is predefined but the per layer rate is determined by the algorithm; and  unstructured weight pruning, where only the proportion of weights pruned is predefined. Their results suggest that for structured and automatic pruning, random initialization is equally (if not more) effective.  However, for unstructured networks, random initialization can achieve similar results on small data sets but for large scale data such as ImageNet, fine-tuning produces better results. 

At first sight, their findings contradict the Lottery Hypothesis.  However, in a follow up study,  \citet{Frankle2019} acknowledge that setting the weights of pruned networks to their initial values does not work well on larger networks and suggest that methods for retraining from random initializations do not work well either, except for moderate levels of pruning (up to 30\%). They therefore propose setting the weights to those obtained in a later iteration of training, which they then demonstrate to be beneficial in identifying good initialization of weights for larger scale problems such as ImageNet.  

The above studies focus on empirical evaluations of networks trained and used on the same data sets, and primarily on image processing classification tasks. \citet{Morcos2019} explore a number of other interesting questions:  
\begin{itemize}
\item Are the lotteries found for one image classification task transferable to other tasks?   
\item Are lotteries observable in other tasks (such as natural language processing), and architectures?  
\item Are they transferable across different  optimizers?
\end{itemize}

To explore these questions, they carry out experiments with VGG19 and ResNet50 using six data sets (Fashion-MNIST, SVHN, CIFAR10, CIFAR100, ImageNet, Places365), in which the lotteries (i.e.,  subnetworks with initializations) identified for one task are used for another task.  Their experiments use iterative magnitude based pruning, selecting 20\% of the weights over all the layers, and with late setting of weights (as proposed in~\citep{Frankle2019a}). The results are interesting:  in general, winning initializations carry across similar image processing tasks and winning tickets from larger scale tasks were more transferable than the tickets from the smaller scale tasks. In some cases, for example, the use of VGG19 on the Fashion-MNIST data, the winning tickets obtained from the use of VGG19 on the larger data sets (CIFAR100, ImageNet) performed better than those obtained directly from the Fashion-MNIST data.   

\citet{Hubens2020} carry out empirical trials that confirm similar results on the size of  the pruned networks. They show that when a network is trained on a larger data set, such as ImageNet, and transferred and fine-tuned for a different task, pruning can result in a smaller network than if it was trained from scratch on the new task. 

\citet{Morcos2019} carry out experiments in which lottery tickets are identified using one optimizer,  ADAM (adaptive moment estimation), and then utilise a different optimizer, SGD (Stochastic Gradient Descent) with momentum, and vice versa on the CIFAR10 data. Their results suggest that, in general, winning tickets are optimizer independent.

To test if the lottery hypothesis holds in other types of problems,~\citet{Yu2019} carry out experiments on natural language processing (NLP)  and control tasks in games. For NLP, they utilise LSTMs for the Wikitext-2 data~\citep{Merity2017} and Transformer models for translating news in English to German~\citep{Vaswani2017}. The experiments were carried out with 20 rounds of iterative pruning  and with one-shot pruning. A pruning rate of 20\% was used and following pruning, weights were reset to those learned during a later round of training. For control tasks, they utilise Reinforcement Learning (RL) and carry out experiments on fully connected networks used for 3 OpenAI Gym environments~\citep{Brockman2016} and 9 Atari games that utilise convolutional networks~\citep{Bellemare2015}.

From their results on NLP and the RL control tasks, they conclude that both iterative pruning and late setting of weights are superior in comparison to random initialization of pruned networks, with iterative pruning being essential when a significant number of weights (i.e., more than two-thirds) are pruned.  For the Atari games, the results varied:  in one case, it led to improvements over the original network (Berzerk game) while in another, an initial improvement was followed by a significant drop in accuracy as the amount of pruning increased (Space Invaders game). In other cases, pruning resulted in a reduction in performance (e.g., Assault game).  Thus in summary, ~\citet{Yu2019} provide some evidence that the lottery hypothesis holds for NLP tasks and for some control tasks that utilise RL. 

%Given the above evidence that a magnitude based iterative pruning scheme can identify good initializations, one is left wondering if alternative initializations or methods %for identifying subnetworks can also be effective?  \citet{Zhou2019} shed some light on this question. They experiment with several different options on the MNIST and %CIFAR10 data and conclude that pruning weights based on the difference between their absolute initial and final values is also an effective criteria, and that, when it %comes to initializing the weights, the essential requirement is to ensure that they have the same sign.  Thus, it is interesting  that alternative criteria for pruning and %initialization can also be effective, suggesting that our understanding of lotteries is not yet complete and more research is needed.

\subsection{Pruning Feature Maps and Filters}
\label{subsectionc:featuremaps}

Although the kind of methods described in Subsection~\ref{subsectiona:netpruning} result in fewer weights,  they require specialist libraries or hardware for processing the resulting sparse weight matrices \citep{Li2017, Denil2013, Ayinde2019}.
In contrast, pruning at higher levels of granularity, such as pruning filters and channels benefits from the optimizations already available in many current toolkits. This has led to a number of  methods for pruning feature maps and filters which are summarized in this section.

\begin{figure}

\begin{center}
\includegraphics[scale=0.8]{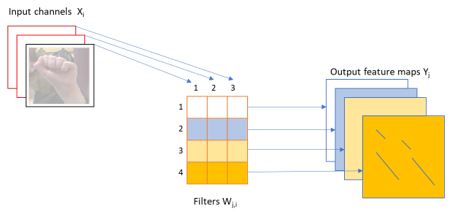}
\end{center}

\caption{Fig. 3.  Illustration of how the feature maps are computed, where $W_{j,i}$  are the k x k filters used on the input channels $X_{i}$ to obtain output feature maps $Y_{j}$
}
\label{figure:illustration_fms}

 \end{figure}
To appreciate the intuition and notation behind these methods,  it is worth bearing in mind how filters are applied to the input channels to produce the output feature maps. Figure~\ref{figure:illustration_fms} illustrates the process,  showing how an image with 3 channels is taken as input and convolved with the filters to produce the 4 output feature maps.
 Given the visualisation offered by Figure~\ref{figure:illustration_fms}, how can one best prune the filters and channels?  The survey revealed three main directions of research:
\begin{enumerate}
\item {\bf Data dependent channel pruning methods}, which are based on the view that when different inputs are presented, the output channels (i.e., feature maps) should vary given they are meant to detect discriminative features.  Subsections~\ref{subsubsectionc1:variancechannels} and~\ref{subsubsectionc2:entropybased} describe the methods that adopt this view.

\item {\bf Data independent pruning methods}, that use properties of the filters and output channels, such as the proportion of zeros present, to decide which filters and channels should be pruned.  Subsections~\ref{subsubsectionc3:apoz} to \ref{subsubsectionc5:aofp} describe methods that take this direction. 
\item {\bf Optimization based channel approximation pruning methods}, that use optimization methods to recreate the filters to approximate the output feature maps. Pruning methods that typify this approach are described in Subsections \ref{subsubsectionc5:aofp} and \ref{subsubsectionc7:lasso}.

\end{enumerate}

\subsubsection{Pruning based on variance of channels and filters}
\label{subsubsectionc1:variancechannels}

\citet{Polyak2015} propose two methods for pruning channels:  Inbound pruning, which aims to reduce the number of channels incoming to a filter and Reduce and Reuse pruning, which aims to reduce the number of output channels.

The idea behind Inbound pruning is to assess the extent to which an input channel's contribution to producing an output feature map varies with different examples.  This assessment is done by applying the network to a sample of the images and then using the variance in a feature map as a measure of its contribution.  

More formally, given $W_{j,i}$, the $j_{th}$ filter for the $i^{th}$  input channel, and  $X_{i}^{p}$, the input from the $i^{th}$ channel for the $p^{th}$  example, the contribution to the $j^{th}$  output feature map, $Y_{j,i}^{p}$ is defined by:
\begin{equation}
  Y_{j, i}^{p}=\left\|W_{j, i} \cdot X_{i}^{p}\right\|_{F}
\end{equation}
Given this definition, the measure used to assess the variation in its contribution, $\sigma_{j,i}^{2}$ from the $N$  samples is:		
\begin{equation}
\sigma_{j, i}^{2} = {var}\left(\left\{Y_{j, i}^{p} \mid p=1 \ldots N\right\}\right)
\end{equation}
Inbound pruning uses this measure to rank the filters $W_{j,i}$ and removes any that fall below a specified threshold.

The Reduce and Reuse pruning method focuses on assessing the variations in the output feature maps when different samples are presented.  That is, the method first computes the variations in the output feature maps $\sigma_{j}^{2}$  using:

\begin{equation}
\sigma_{j}^{2}={var}\left(\left\|\sum_{i=1}^{m} Y_{j, i}^{p}\right\|_{F} \mid p=1 \ldots N\right)
\end{equation}
Where $m$ is the number of input channels and $N$ is the number of samples. 
Reduce and Reuse then uses this measure to retain a proportion of the output feature maps and corresponding filters that results in the greatest variation.

Removal of an output feature map is problematic given it is expected as an input channel in the next layer. To overcome this, they approximate a removed channel using the other channels.  That is, if $Y_{i}$,  $Y_{i}^{\prime}$ are the outputs of a layer before and after pruning a layer respectively, the aim is to find a matrix A such that:

\begin{equation}
\min _{A} \sum_{i}\left\|Y_{i}-A Y_{i}^{\prime}\right\|_{2}^{2}
\end{equation}

The matrix $A$ is then included as an additional convolutional layer of 1x1 filters along the lines proposed by~\citet{Lin2014}. 

\citet{Polyak2015} evaluate the above approach on the Scratch network, using the CASIA-WebFace and the Labeled Faces in the Wild (LFW) data sets. They utilise layer by layer pruning, where each layer is pruned, and the network fine-tuned before moving on to the next layer. They experiment with their two pruning methods individually and in combination, and compare the results  with the use of random pruning, a low rank approximation method \citep{Zhang2016} and Fitnets, a method that uses the Knowledge Distillation approach to learn smaller networks \citep{Romero2014}. In the experiments with the Inbound pruning method, they prune channels where $\sigma_{j,i}^{2}$ is below a given threshold, selected such that the overall accuracy is maintained above 84\%.  For the experiments with the Reduce and Reuse method, they try different levels of pruning:  50\%, 75\%, and 90\% for the earlier layers followed by 50\% for the later layers.  The adoption of a lower pruning rate for the later layers follows an observation that heavy pruning of the later layers results in a marked reduction in accuracy.  

The results from their experiments show that: (i) the variance based method is more effective than use of random pruning,  (ii) the use of fine-tuning does help in recovering accuracy, especially in the later layers, (iii) their methods result in greater compression than use of  a low rank method and the use of Fitnets  when applied to the Scratch network.

\subsubsection{Entropy-based channel pruning}

\label{subsubsectionc2:entropybased}
Instead of the variance, \citet{Luo2017a} propose an entropy-based metric to evaluate the importance of each filter. In their filter pruning method, if a feature map contains less information, its corresponding filter is considered less important, and could be pruned. To compute the entropy value of a particular feature map, they first sample the data and obtain a set of feature maps for each filter.  Each feature map is reduced to a point measure using a global average pooling method, and the set of measures associated with each filter are discretized into q groups. The entropy of a filter, $H_{j}$  is then used to assess the discriminative power of a filter~\citep{Luo2017a}:
\begin{equation}
H_{j}=\sum_{i=1}^{q} P_{i} * \log \left(P_{i}\right)
\end{equation}
Where $P_{i}$  is the probability of an example being in group $i$.

They explore both one-shot pruning followed by fine-tuning and layer wise pruning in which they fine-tune with just one or two epochs of learning immediately after pruning a layer.  Their layer wise strategy is an interesting compromise between fully fine-tuning after pruning each layer, which can be computationally expensive, and only fine-tuning at the end, which can fail to take account of the knock-on effects of pruning previous layers.

They evaluate the merits of the entropy-based method by applying it to VGG16 and ResNet-50 on the ImageNet data.  For VGG16, they focus on the first 10 layers and, also replace the fully connected layers by use of average pooling to obtain further reductions. They compare their results on VGG16 with those obtained by the magnitude based pruning method and APoZ method (c/f Subsection~\ref{subsubsectionc3:apoz}). Their results suggest that: (i) the entropy-based method achieves more than a 16 fold compression, though this is at the expense of a 1.56\% reduction in accuracy,  (ii) use of magnitude pruning  results in a 13 fold compression, and (iii) APoZ results in a 2.7 fold compression.  However, it should be noted that the higher compression rate achieved by the use of entropy includes the reduction due to the replacement of the fully connected layers by average pooling, without which the use of the entropy-based  method leads to a lower compression rate than APoZ (Table 3 in  \cite{Luo2017a}).

\subsubsection{APoZ: Network trimming based on zeros in a channel}
\label{subsubsectionc3:apoz}

In contrast to the use of  samples of data to compute the variance of a feature map or its entropy, \citet{Hu2016}, suggest a direct method that is based on the view that the number of zeros in an output feature map is indicative of its redundancy. Based on this view, they propose a method that uses the average number of zero activations (APoZ) in a feature map (after the ReLU) as a measure of  the weakness of a filter that generates the feature map.

Their experiments are with LeNet5 on MNIST and VGG16 on ImageNet and aimed at first finding the most appropriate layers to prune and  then to iteratively prune these layers in a bespoke way that maintains or improves accuracy. Following pruning, they experiment with both retraining from scratch and  fine-tuning the weights and prefer the latter given better results. 

For LeNet-5, they observe that most of the parameters (over 90\%) are in the 2nd convolution layer and the first fully connected layer and hence they focus on pruning these two layers in four iterations of pruning and fine-tuning, resulting in the size of the convolutional layer reducing  from 50 to 24 filters  and the number of neurons in the fully connected  layer reducing from 500 to 252.  Overall, this represents a compression rate of 3.85.

For VGG16, they also focus on one convolutional layer that has 512 filters and a fully connected layer with 4096 nodes.  After 6 iterations, they reduce these to 390 filters and 1513 nodes, achieving a compression rate of 2.59.

\subsubsection{Pruning small filters and filter sketching}
\label{subsubsectionc4:smallfilters}
\citet{Li2017} extend the idea of magnitude pruning of weights to filters by proposing the removal of filters that have the smallest absolute sum among the filters in a layer. That is, if the filters for producing the $j^{th}$ feature map are  
$W_{j,i} \in \R^{k x k}$ and $m$
is the number of input feature maps, then the magnitude of the  $j^{th}$  filter is defined by:
\begin{equation}
s_{j}=\sum_{i=1}^{m}\left\|W_{j, i}\right\|_{1}
\end{equation}
Once the $s_{j}$  are computed, a proportion of the smallest filters together with their associated feature maps and filters in the next layer are removed. After a layer is pruned, the network is fine-tuned, and pruning is continued layer by layer.  

To test this approach, they carry out experiments on VGG16 and ResNet56 \& 110 on CIFAR10 and ResNet34 on ImageNet.  By analyzing the sensitivity of the layers through experimentation, they determine appropriate pruning ratios for each layer that would not compromise accuracy significantly. Overall, for VGG16, they are able to prune the parameters by 64\%. A significant proportion of this pruning is in layers 8 to 13 which consist of the smaller filters (2x2 and 4x4), which they notice can be pruned by 50\% without reducing accuracy.  The level of pruning for the other networks is more modest, with the best pruning rate for ResNet-56 and ResNet110 on CIFAR10  being 3.7\%  and 32.4\% respectively, and for ResNet-34 on ImageNet being 10.8\%.

They also compare their approach with the variance-based method (Subsection~\ref{subsubsectionc1:variancechannels}) and conclude that use of the above measure over filters performs at least as well but without the additional need to compute the feature maps via samples of the data.

A recent method known as filter sketch also aims to reduce the number of filters without the need to sample examples~\citep{Lin2020}.   The key idea in filter sketching is to minimize the difference between the co-variances of the original set of filters and the reduced set. Although this can be done using optimization methods, \citet{Lin2020} utilise a greedy algorithm known as  Frequent Direction \citep{Liberty2013} which is more efficient.   

\citet{Lin2020} evaluate the filter sketch method on GoogleNet, ResNet56 and ResNet110 using the CIFAR10 data, and on ResNet50 with the ImageNet data.  The results show that it performs well relative to the method for pruning small filters and a method that uses optimization to  prune channels (c/f Subsection~\ref{subsubsectionc7:lasso}) in terms of reducing the number of parameters without a significant loss in accuracy.  

\subsubsection{Pruning filters based on geometric median}
\label{subsubsectionc5:geometricmedian}

\citet{He2019} point out that pruning based on the magnitude of filters assumes that there are some small filters and that the spread of magnitude is wide enough to adequately distinguish those filters that contribute from those do not contribute.  So, for example, if most of the weights are small, one could end up removing a significant number of filters and if most of the filters have large values, no filters would be removed, even though there may be filters that are relatively small.  Hence, they propose a method based on the view that the geometric median of the filters shares most of the information common in the other filters and hence a filter that is close to it can be covered by the other filters if deleted.  Computing the geometric median can be time-consuming, so they approximate its computation by assuming that one of the filters will be the geometric mean.  Their pruning strategy is to prune and fine-tune repeatedly using a fixed pruning factor for all layers. 

They carry out an evaluation with respect to several methods including pruning small filters~\citep{Li2017}, ThiNet~\citep{Luo2017}, Soft filter pruning~\citep{He2018},  and  NISP~\citep{Yu2018}.  These methods are evaluated on ResNets trained on the CIFAR10 and ImageNet data, with pruning rates of 30 and 40 percent.   In general, the drop in accuracy is similar across the different methods, though there is a significant reduction in FLOPS when using the geometric median method on ResNet-50 (53.5\%) compared to the other methods (e.g., ThiNet 36.7\%, Soft filter pruning 41\%, NISP 44\%).

\subsubsection{ThiNet and AOFP}
\label{subsubsectionc5:aofp}

 \citet{Luo2017} formulate the pruning task as an optimization problem and propose a system ThiNet in which the objective is to find a subset of input channels that can best approximate the output feature maps. The channels not in the subset and their corresponding filters can then be removed.  Solving the optimization problem is computationally challenging, so ThiNet uses a greedy algorithm  that finds a channel that contributes the least, adds it to the list to be removed, and repeats the process with the remaining channels until the number of channels selected equals the number to be pruned. Once a subset of filters to be retained is identified, their weights are obtained by using least squares to find the filters $W$ that minimize~\citep{Luo2017}:
\begin{equation}
\sum_{i=1}^{m}\left(Y_{i}-W^{T} \cdot X_{i}\right)^{2}
\end{equation}
Where $Y_i$  are the $m$ sampled points in the output channels and $X_i$ their corresponding input channels.

They evaluate their approach in two sets of experiments.  In the first, they adapt VGG16, replacing the fully connected layers by global average pooling (GAP) layers, apply it to the UCSD-Birds data and then prune it using ThiNet, APoZ and the small filters method. Their results show there is less degradation in accuracy with ThiNet than ApoZ, which in turn, is better than the small filters method.

In their second set of experiments, they utilise VGG16 and ResNet50 trained on the ImageNet data.  For VGG16, their procedure involves pruning a layer and then minor fine-tuning with one epoch of training with an additional epoch at the end of each group of convolutional layers and a further 12 epochs of fine-tuning after the final layer.   With the use of GAP, ThiNet, reduces the number of parameters by about 94\% at the expense of a 1\% reduction in Top-1 accuracy. For ResNet, ThiNet is applied on the first two convolutional layers of each residual block, keeping the output dimensions of the blocks the same.  After pruning each layer, one epoch of fine-tuning is performed, and 9 epochs are used for fine-tuning at the end.  The results show that ThiNet is able to halve the number of parameters with a 1.87\% loss in Top-1 accuracy. 

\citet{Ding2019c} propose a similar method to ThiNet, called Approximated Oracle Filter Pruning (AOFP),  which aims to identify the subset of filters, which if removed, will have the least effect on the feature maps in the next layer.  However, whereas, the search procedure adopted in ThiNet uses a greedy bottom up approach, AOFP adopts a top-down binary search in which half of the filters in a layer are randomly selected and set to be pruned.  The effect of removing these filters on the feature map produced in the next layer is measured and recorded against each filter that is set as pruned.   This process is repeated for different random selections, and the average effect per filter used as an indication of the effect of removing a filter.   The top 50\% of the filters that would result in the worst effect if removed are retained and the process repeated unless this would result in an unacceptable reduction in accuracy. In comparison to ThiNet, and other  methods, AOFP does not require the rate of pruning to be fixed in advance of pruning a layer.

AOFP is evaluated by pruning AlexNet, VGG and ResNet trained on the CIFAR10 and ImageNet data.  They compare AOFP with several methods including: ThinNet, Network Slimming~\citep{Liu2017}, Pruning using Agents~\citep{Huang2018a},  Online Filter Weakening~\citep{Zhou2018}, NISP~\citep{Yu2018},  Optimizing Channel Pruning~\citep{He2017}, Structured Probabilistic Pruning~\citep{Wang2017}, Autopruner ~\citep{Luo2018}, and ISTA~\citep{Ye2018}, with their results showing that AOFP is capable of greater reductions in FLOPS without compromising accuracy.  

\subsubsection{Optimizing channel selection with LASSO regression}
\label{subsubsectionc7:lasso}

\citet{He2017} also formulate channel selection as an optimization problem. Given a channel $Y$  obtained by applying a filter  $W_{i}$ to  $m$  input channels $X_{i}$ :

\begin{equation}
Y=\sum_{i=1}^{m} X_{i} W_{i}^{T}
\end{equation}
They define the task  as one to optimize:
\begin{equation}
	\begin{aligned}
	& \arg \min _{\beta, W} \frac{1}{2}\left\|Y-\sum_{i=1}^{c} \beta_{i} X_{i} W_{i}^{T}\right\|_{F}^{2}\\
	&\text{subject to } \|\beta\|_{0} \leq p  
	\end{aligned}
\end{equation}
Where $p$ indicates the number of channels retained and $\beta_{i} \in \{0,1\} $ indicates the retention or removal of a channel.

In contrast to ThiNet, which adopts a greedy heuristic to solve this optimization problem, \citet{He2017} relax the problem from $L_{0}$ to $L_{1}$ regularization and utilise LASSO regression to solve :
\begin{equation}
\begin{aligned}
& \arg \min _{\beta, W} \frac{1}{2}\left\|Y-\sum_{i=1}^{c} \beta_{i} X_{i} W_{i}^{T}\right\|_{F}^{2}+\lambda\|\beta\|_{1} \\
& \text{subject to }  \|\beta\|_{0} \leq p 
\end{aligned}
\end{equation}
Following the selection of the channels they  utilise least squares to obtain the revised weights in a manner similar to the approach adopted in ThiNet.

They carry out empirical evaluations on VGG16, ResNet50 and a version of the Xception network, trained on the CIFAR10 and ImageNet data.  They also explore the extent to which the pruned models can be used for transfer learning by using them for the PASCAL VOC 2007 object detection task.  

In their first set of experiments,  they evaluate their method on single layers of VGG16 trained on CIFAR10 without any fine-tuning, and show that their algorithm maintains Top-5 accuracy better than the method of pruning small filters.  They also include results from a na\"ive method that selects the first k feature maps and show that, for some layers (e.g. conv3\_3 in VGG16), this sometimes performs better than the method of  pruning small filters, highlighting a potential weakness of magnitude-based pruning.

In a second set of experiments, with VGG16 on CIFAR10, they apply their method on the full network, using bespoke pruning ratios for the layers and fine-tuning to achieve 2, 4 and 5 fold improvements in run-time, but resulting in drops of  Top-5 accuracy of 0\%,1\%, and 1.7\% respectively.  In comparison, the method for pruning small filters results in larger drops of 0.8\%, 8.6\% and 14.6\%.

Their experiments on ResNet50 adopt bespoke pruning rates per layer, retaining 70\% of layers that are very sensitive to pruning, and 30\% of the less sensitive layers.  The  Top-5 accuracy results on ImageNet show a two-fold improvement in run-time at the expense of a 1.4\% drop in accuracy compared to a baseline accuracy of 92.2\%, while the results on the Xception network show a drop of 1\% in accuracy from a baseline of 92.8\%.

The experiment on using a pruned version of a VGG16 model on  the PASCAL VOC 2007 object detection benchmark task results in a 2-fold increase in speed with a 0.4\% drop in average precision. 

\section{PRUNING BASED ON SIMILARITY AND CLUSTERING}
\label{section:clustering}

%An alternative approach to pruning is to use similarity and clustering to identify and remove parameters that unnecessarily duplicate% %functionality\citep{Sussmann1992,RoyChowdhury2017,Han2016b,Son2018,Li2019sv}, and this section summarises two recent studies.  

Given that neural networks can be over-parametrised, it is plausible that there could be duplicate weights or filters that  perform similar functions and can be removed without impacting accuracy \citep{Sussmann1992,RoyChowdhury2017,Han2016b,Son2018,Li2019sv}.

\citet{RoyChowdhury2017} explore this hypothesis by using the inner product of two filters (or weight matrices) as a measure of similarity.  Their pruning algorithm involves grouping filters that are similar and then replacing each group of filters by their mean filter. 
They carry out experiments with both a multilayer perceptron (MLP) and a CNN for the CIFAR10 data. The MLP has three layers: the first two are fully connected layers and the third is a softmax layer with 10 nodes representing the class for CIFAR10. The CNN has two convolution layers, each followed by a ReLU, and a 2x2 max pooling layer.  The convolutional layers are followed by two fully connected layers to perform the classification.  In both cases, the first layer is varied with 100, 500 and 1000 units (nodes or filters)  to explore the effects of increasing over parametrisation.		 
Their main finding is that there is a much greater propensity for similar weights/filters to occur in MLPs than  in CNNs.  As a consequence, there is a greater opportunity for using similarity as a basis for pruning MLPs than for pruning CNNs. Nevertheless, their results suggest that a similarity based pruning algorithm is better at retaining accuracy  than using the small filters method.

\citet{Ayinde2019} also develop a method that uses clustering to identify similar filters.  They too adopt the inner product as a measure of similarity, but use an agglomerative hierarchical clustering method to group similar filters and replace the filters by randomly selecting one filter from each cluster. 
They carry out various experiments with VGG16 on CIFAR10 and ResNet34 on ImageNet.     
For the trial on VGG16 with the CIFAR10 data, they show that, once an optimal value for the threshold for similarity is determined, their method achieves both a better pruning rate and accuracy than  other methods, including pruning of small filters, Network Slimming, a method that uses regularization to identify weak channels~\citep{Liu2017},  and try-and-learn, a method that uses sensitivity analysis (c/f Section \ref{section:sensitivity}).

\section{SENSITIVITY ANALYSIS METHODS}
\label{section:sensitivity}

The primary goal of pruning is to remove weights, filters and channels that have least effect on the accuracy of a model. The magnitude and similarity based methods described above address this goal implicitly by using properties of weights, filters and channels that can affect accuracy.  In contrast, this section presents methods that use sensitivity analysis to model the effect of perturbing and removing  weights, filters and channels on the loss function.  

%Subsection~\ref{subsubsection:pruningchannels} begins with a description of Skeletonization, one of the first studies to propose the use of sensitivity analysis, and is followed by a summary of methods that assess the importance of channels. Subsections~\ref{subsubsection:obd}
% to~\ref{subsubsection:secondorder} present the development of  a line of research that approximates the effect of perturbing the weights on the loss function using the %Taylor Series, from the earliest work which developed methods for MLPs to the more recent research that is aimed at pruning CNNs.

Subsection~\ref{subsubsection:pruningchannels} describes methods that assess the importance of channels and Subsections~\ref{subsubsection:obd}
to~\ref{subsubsection:secondorder} present the development of  a line of research that approximates the effect of perturbing the weights on the loss function using the Taylor Series, from the earliest work which developed methods for MLPs to the more recent research on methods for pruning CNNs.

\subsection{Pruning by assessing the importance of nodes and channels}
\label{subsubsection:pruningchannels}
 
Skeletonization, a method proposed  by~\citet{Mozer1988}, was one of the earliest approaches to pruning neural networks. To calculate the effect of removing nodes, Skeletonization introduced the notion of attentional strength to denote the importance of nodes when computing  activations.  Given the attentional strengths of the nodes, $\alpha_{i}$,  the  output $y_{j}$  from node $j$, is defined by:
\begin{equation}
y_{j}=f\left(\sum_{i} w_{j i} \alpha_{i} y_{i}\right)
\end{equation}
Where $f$ is assumed to be the sigmoid function.
The importance of a node  $\rho_{i}$  is then defined in terms of the difference in loss when $\alpha_{i}$ is set to zero and when it is set to one and  can be approximated by the derivative of the loss with respect to the attentional strength $\alpha_{i}$:
\begin{equation}
\rho_{i}=\mathcal{L}_{\alpha_{i}=0}-\mathcal{L}_{\alpha_{i}=1} \approx-\left.\frac{\partial \mathcal{L}}{\partial \alpha_{i}}\right|_{\alpha_{i}=1}
\end{equation}

%simplified description
%Although \citet{Mozer1988} attempted to utilise the quadratic loss function for the purposes of training, they concluded that it did not perform well when computing  %$\rho_{i}$   given that the outputs and targets were close following training.  They therefore adopted a linear loss function when computing $\rho_{i}$  :
%\begin{equation}
%\sum_{p, j}\left|t_{j}^{p}-y_{j}^{p}\right|
%\end{equation}
%Where $t_{j}^{p}$ are the targets for output unit $j$  for example (pattern) $p$  and  $y_{j}^{p}$  are the corresponding outputs from the network.

Through experimentation, \citet{Mozer1988} found that the linear loss  worked better a than the quadratic loss because the difference between the outputs and targets was small following training.
In addition, they noticed that the $\partial \mathcal{L}(t)$/$\partial \alpha_{i}$  were not stable with time, so they used a weighted average measure to compure the importance $\widehat{\rho_{i}}$:
\begin{equation}
\widehat{\rho_{i}}(\mathrm{t}+1)=0.8 \widehat{\rho_{i}}(\mathrm{t})+0.2 \frac{\partial \mathcal{L}(t)}{\partial \alpha_{i}}
\end{equation}
\citet{Mozer1988} present a number of small but very interesting experiments.  These include generating examples where the output is correlated to four inputs, A,B,C, and D,  with full correlation on A and reducing to no correlation with D.  They provide this as input to a network with one hidden node and following training they observe that the weights from the inputs to the hidden node follow the correlations, although the relevance measure only shows input node A as important, providing some reassurance that the  measure is different from the weights. In another example, they develop a network to model a 4-bit multiplexor network, which has 4 bits as input and two bits to control which of the 4 bits is output.  They try two network configurations:  in the first, they utilise 4 hidden nodes and in the second they utilise 8 hidden nodes and use skeletonization to reduce its size to 4 hidden nodes.  When limiting training to 1000 epochs, they find that starting with 4 hidden nodes initially, results in failure to converge in 17\% of the cases, while beginning with 8 hidden layer nodes followed by skeletonization converges in all the cases and, also retains accuracy. This appears to be one of the first demonstrations that, to begin, it may be necessary to overparameterize a network in order to find winning lotteries. 

This idea of assessing the importance of nodes has been extended to channels by two methods, namely Network Slimming  \citep{Liu2017} and Sparse Structure Selection (SSS)~\citep{Huang2018}, that learn a measure of importance as part of the training process.  Both utilise a parameter $\gamma$  for each channel (analogous to the attentional strength) which scales the output of a channel.  Given a loss function $\mathcal{L}$, the new loss  $\mathcal{L}^{\prime}$ is defined with an additional regularization term over the scaling factors $\gamma$:
\begin{equation}
\mathcal{L}^{\prime}=\mathcal{L}+\lambda \sum_{u} g(\gamma)
\end{equation}
Where the function $g$ is selected as the $L_1$ norm to reduce $\gamma$ towards zero  (as in Lasso regression).

The two methods differ in the way they implement the training process aimed at minimizing  $\mathcal{L}^{\prime}$ with Network Slimming taking advantage of the use of batch normalization layers that are sometimes present following convolutional layers while SSS implements a more general process that does not assume the presence of batch layers, and allows use of scaling factors for blocks (such as residual and inception blocks) that can enable reduction of the depth of a network.

\citet{Huang2018} experiment with SSS  on the CIFAR-10, CIFAR-100, and ImageNet data on VGG16, and ResNet. 
For CIFAR10, SSS is able to reduce the number of parameters in VGG16 by 30\% without loss of accuracy.  For ResNet-164, it is able to achieve  a 2.5 times speedup at the cost of a 2\% loss in accuracy for CIFAR-10 and CIFAR-100. 

For VGG16 on ImageNet, SSS is able to reduce the FLOPs by about 75\%, though parameter reduction is minimal, which is consistent with other methods given the large number of parameters in the fully connected layers in VGG16.  On ResNet50, SSS achieves  a 15\% reduction in FLOPs at a cost of a 0.1\% reduction in Top-1 accuracy.

\subsection{Pruning Weights with OBD and OBS}
\label{subsubsection:obd}

Several studies  utilise the Taylor series to approximate the effect of weight perturbations on the loss function \citep{LeCun1990,Hassibi1993,Wang2019}. Given the change in weights $\Delta W$, a Taylor Series approximation of the change in loss  $\Delta \mathcal{L}$ can be stated as~\citep{Bishop2006}:
\begin{equation}
\Delta \mathcal{L}=\frac{\partial \mathcal{L}^{T}}{\partial W} \Delta W+\frac{1}{2} \Delta W^{T} H \Delta W+O\left(\|\delta W\|^{3}\right)
\end{equation}
Where $H$ is a Hessian matrix whose elements are the second order derivatives of the loss with respect to the weights:
\begin{equation}
H_{i j}=\frac{\partial^{2} \mathcal{L}}{\partial w_{i} \partial w_{j}}
\end{equation}

Most methods that adopt this approximation assume that the third order term is negligible. In Optimal Brain Damage (OBD), \citet{LeCun1990},  also  assume that the first order term can be ignored given that the network will have been trained to achieve a local minima, resulting in a simplified quadratic approximation:
\begin{equation}
\Delta \mathcal{L}=\frac{1}{2} \Delta W^{T} H \Delta W
\end{equation}
Given the large number of weights, computing the Hessian is computationally expensive, so they also assume that the change in loss can be approximated by the diagonal elements of the Hessian,  resulting in the following measure of the saliency $s_{k}$   of a weight $w_{k}$:
\begin{equation}
s_{k}=\frac{H_{k k} w_{k}^{2}}{2}
\end{equation}
Where the second order derivatives, $H_{k k}$ are computed in a manner similar to the way the gradient is computed in backpropagation.   

\citet{Hassibi1993} argue that ignoring the non-diagonal elements of a Hessian is a strong assumption, and propose an alternative pruning method, called Optimal Brain Surgeon (OBS), that aims to take account of all the elements of a Hessian ~\citep{Hassibi1993,Hassibi1993a}.  

Using a unit vector, $e_m$,  to denote the selection of the $m_{th}$  weight as the one to be pruned, OBS reformulates pruning as a constraint-based optimization task: 
\begin{equation}
\begin{aligned}
& \min _{\delta w}\left\{\frac{1}{2} \delta w^{T} \cdot H \cdot \delta w\right\}  \\
& \text{ subject to } e_{m}^{T} . \delta w+\delta w_{m}=0
\end{aligned}
\end{equation}
Formulating this with a Lagrangian multiplier, $\lambda$, the task is to minimize:
\begin{equation}
\frac{1}{2} \delta w^{T} . H . \delta w+\lambda\left(e_{m}^{T} . \delta w+\delta w_{m}\right)
\end{equation}
By taking derivatives and utilizing the above constraint, they  show the saliency, $s_{k}$ of weight $w_{k}$   can be computed using:
\begin{equation}
s_{k}=\frac{1}{2} \frac{w_{k}^{2}}{\left[H^{-1}\right]_{k, k}}
\end{equation}

They show that on the XOR problem, modelled using a MLP network  with 2 inputs, 2 hidden layer nodes and one output,  OBS is better at detecting the correct weights to delete than OBD or magnitude pruning.  They also show that OBS is able to significantly reduce the number of weights  required for neural networks trained on the Monk problems~\citep{Thrun1991} and for NetTalk~\citep{Sejnowski1987}, one of the classical applications of neural networks,  it is able to reduce the number of weights required from 18000 to 1560.  

\subsection{Pruning  Feature Maps with First-Order Taylor Approximations}
\label{subsubsection:taylor}
The methods described in Subsection~\ref{subsubsection:obd} focus on the effect of removing weights in a fully connected network. \citet{Molchanov2016a} introduce a method that uses the Taylor series to approximate what happens if a feature map is removed. In contrast to OBD and OBS, which assume that the first order term can be ignored, they adopt  a first order approximation, ignoring the higher order terms, primarily on grounds of computational complexity. Using a first order approximation seems odd given the convincing argument for ignoring these terms; however  they argue that although the first order gradient tends to zero, the expected value of the change in loss is proportional to the variance, which is not zero and is a measure of the stability as a local solution is reached.  Given a feature map with $N$  elements $Y_{i,j}$, the first order approximation using the Taylor series leads to the following measure of the absolute change in loss~\citep{Molchanov2016a}:
\begin{equation}
\Delta \mathcal{L}(Y)=\left|\frac{1}{N} \sum_{i, j} \frac{\partial \mathcal{L}}{\partial Y_{i, j}} Y_{i, j}\right|
\end{equation}
The scale of this measure will vary in different layers, and they  therefore apply $L_{2}$  normalization within each layer.
The pruning process they adopt involves selecting a feature map using the measure, pruning it, and then fine-tuning the network before repeating the process until a stopping condition, that takes account of the need to reduce the number of FLOPs while maintaining accuracy, is met. 
Their experiments reveal several interesting findings:
\begin{enumerate}
\item From experiments on VGG16 and AlexNet on the UCSD-Birds and Oxford-Flowers data, they show that the features maps selected by their criteria correlate significantly more closely to those selected by an oracle method than OBD and APoZ.  On the ImageNet data, they find that OBD correlates best when AlexNet is used.
\item In experiments on transfer learning, where they fine-tune VGG16 on the UCSD-Birds data, they present results showing that their method performs better than APoZ and OBD as the number of parameters pruned increases. In an experiment in which AlexNet is fine-tuned for the Oxford Flowers data, they show that both their method and OBD perform better than APoZ.
\item In a striking example of the potential benefits of pruning, they demonstrate their method on a network for recognizing hand gestures that requires over 37 GFLOPs for a single inference but only requires 3 GFLOPs after pruning, all be it with a 2.6\% reduction in accuracy.    
\end{enumerate}

In a follow up publication, \citet{Molchanov2019}  acknowledge some limitations of the above approach, namely that assuming that all layers have the same importance does not work for skip connections (used in the ResNet architecture) and that assessing the impact of changes in feature maps leads to increases in memory requirements.  They therefore propose an alternative formulation, also using a Taylor series approximation, but based on estimating the squared loss due to the removal of the $m_{th}$  parameter:

\begin{equation}
	\label{eqn23}
\left(\Delta \mathcal{L}_{m}\right)^{2}=\left(g_{m} w_{m}+\frac{1}{2} w_{m} H_{m} W\right)^{2}\\
\end{equation}
 Where $g_{m}$ is the first order gradient and $H_{m}$ is the  $m^{\text {th }}$  row of the Hessian matrix. 
The measure of importance of a filter is then obtained by summing the contributions due to each parameter in a filter.

The pruning algorithm employed proceeds as follows.  In each epoch, they utilise a fixed number of mini-batches to estimate the importance of each filter and then, based on their importance, a predefined number of filters is removed.  The network is then fine-tuned, and the process repeated until a pruning goal, such as the desired number of filters or a limit for an acceptable drop in accuracy is reached.

\citet{Molchanov2019} carry out initial experiments on versions of LeNet and ResNet  on the CIFAR10 data, using both the second and first order approximations (in  equation~\ref{eqn23}) and given that the results from both correlate well with an oracle method, they utilise the first-order measure which is significantly more efficient to compute.  

In experiments with  versions of ResNet, VGG, and  DenseNet on the ImageNet data, they consider the effect of using the measure of importance at points before and after the batch normalization layers, and conclude that the latter option results in greater correlation with an oracle method.  The results from their method show that it works especially well on pruning ResNet-50 and ResNet-34, outperforming the  results from ThiNet and NISP. The reported results for other networks are also impressive, with their method able to prune 76\%  of the parameters in VGG with a 0.19\% loss in accuracy and able to reduce the number of parameters in DenseNet by 43\% at the expense of a 0.29\% reduction in accuracy.

\subsection{Pruning Feature Maps with Second-Order Taylor Approximations}
\label{subsubsection:secondorder}

The first order methods described above assume minimal interaction across channels and filters. This section summarizes recent pruning methods that aim to take account of the effect of the potential dependencies amongst the channels and filters.

In  a method called EigenDamage, that  also utilises the Taylor series approximation, \citet{Wang2019} revisit the assumptions made by OBD and OBS when approximating the Hessian.  To motivate their method, they begin by illustrating that although OBS is better than OBD when pruning one weight at a time, it is not necessarily superior when pruning multiple weights at a time. This is primarily because OBS does not correctly model the effect of  removing multiple weights, especially when they are correlated.  To avoid this problem,  they utilise a Fisher Information Matrix to approximate the Hessian  and then  they utilise a method, proposed by~\citet{Grosse2016}, to represent a Fisher Matrix by a Kronecker Factored Eigenbasis (KFE). This reparameterization allows pruning to be done in a new space in which the Fisher Matrix is approximately diagonal. Pruning can thus be done by first mapping the weights to a KFE space in which they are approximately independent, and then mapping back the results to the original space. 

EigenDamage is evaluated on VGG, and ResNet on the CIFAR10, CIFAR100 and the Tiny-ImageNet data.  Experiments are carried out with one-shot pruning where fine-tuning is performed at the end and with iterative pruning in which fine-tuning is performed after each cycle.  In both cases, the results show that EigenDamage outperforms adapted versions of OBD, OBS and Network Slimming~\citep{Liu2017}.   

\citet{Peng2019}  also utilise a Taylor series approximation to develop a Collaborative Channel Pruning (CCP) method that is based on a measure of the impact of a combination of channels.  Given a mask $\beta$, where $\beta_{i} = 1$, indicates the retention of a channel and $\beta_{i} = 0$,  indicates a channel to be pruned, they formulate the task as one to find the $\beta_{i}$ that minimize the loss $\mathcal{L}$:
\begin{equation}
\mathcal{L}(\beta, W)=\mathcal{L}(W)+\sum_{i=1}^{c_{o}}\left(\beta_{i}-1\right) g_{i}^{T} w_{i}+\frac{1}{2} \sum_{i=1, j=1}^{c_{o}}\left(\beta_{i}-1\right)\left(\beta_{j}-1\right) w_{i}^{T} H_{i, j} w_{j}
\end{equation}
Where  $g_{i}$  are the first order derivatives of the loss with respect to the weights in the $i^{th}$  output channel, 
$H_{i,j}$  are Hessians, and  $c_{o}$  denotes the number of output channels.

By setting $u_{i}=g_{i}^{T} w_{i}$  and  $s_{i, j}=\frac{1}{2} w_{i}^{T} H_{i, j} w_{j}$
the above equation can be written as the following 0-1 quadratic optimization problem \citep{Peng2019}:
\begin{equation}
\begin{aligned}
&\min _{\beta_{i}} \sum_{i=1}^{c_{o}} u_{i}\left(\beta_{i}-1\right)+\sum_{i=1, j=1}^{c_{o}} s_{i, j}\left(\beta_{i}-1\right)\left(\beta_{j}-1\right) \\
&\text { subject to: }\|\beta\|_{0}=p \text { and } \beta_{i} \in\{0,1\}
\end{aligned}
\end{equation}
Where  $p$  denotes the number of channels to be retained in a layer.
They note that the gradients $g_{i}$ and hence $u_{i}$   can be computed in linear time. However, given the complexity of computing the Hessian matrices, they derive first order approximations for  the loss functions, which they adopt when computing  $s_{i,j}$.   To solve the quadratic optimization problem, they relax the constraint to $\beta_{i} \in [0,1]$  and use a quadratic programming method to find the $\beta_{i}$  which are used to select the top $p$  channels to retain.  They apply the optimization process on each layer to obtain the masks $\beta_{i}$,  use these to prune  and  then perform fine-tuning at the end.

An empirical evaluation of CCP is carried out by pruning the ResNet models trained on the CIFAR10 and ImageNet data, and the results compared to several methods including: pruning small filters, ThinNet,  optimizing channel pruning, Soft Filter pruning~\citep{He2018}, NISP~\citep{Yu2018} and  AutoML~\citep{He2018a}.   For CIFAR10, the experiments are carried out with a pruning rate of 35\% and 40\%, and in each case, CCP has a smaller  drop in accuracy (0.04\% and 0.08\% respectively)  than the other methods, with the exception of  the method for pruning small filters which results in  a small improvement in accuracy (0.02\%).  However, the pruning small filters method has a much lower reduction in the FLOPS (27.6\%) in comparison to CCP (52.6\%). The results for ImageNet show, that for similar reductions in FLOPS, CCP has less of a drop in accuracy than the other methods.  

It's worth noting, that like EigenDamage,  CCP is able to obtain good results without the need for an iterative process that uses fine-tuning after pruning each layer.

\section{A Resource for Comparing Published Results}
\label{section:resource}

As the above sections describe, previous studies of pruning report results on varying data sets, architectures and methods that have evolved with time, making comparison of results across the different studies difficult. The survey provides a resource in the form of a pivot table  that  can be used by the community to explore the reported performance of over 50 methods on different architectures and data. Appendix D shows  how many times each combination of data and architecture has been used, indicating  the wide variety of comparisons possible.

\begin{figure}

 \includegraphics[width=6in]{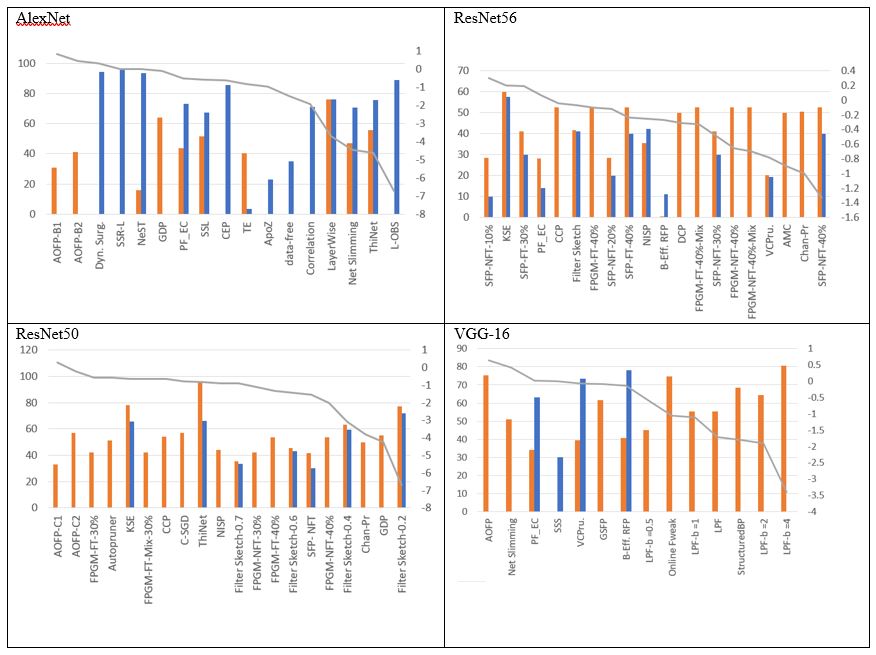}

\caption{Results of pruning AlexNet and, ResNet50 on ImageNet (left column), and  ResNet56 and VGG16 on the CIFAR10 data (right column).  Charts show the percent reduction in parameters where available (blue bars, left axis) and FLOPs (orange bars, left axis), and reduction in baseline Top-1  accuracy (grey line, right axis).  The labels used for the methods are from the primary sources, with suffixes reflecting the variations in pruning rates}

\nocite{Jiang2018,Dong2017,Zhuang2018,Xu2018,Neklyudov2017,Bao2018,Molchanov2016a,Srinivas2015,Sun2016}

\label{figure:fig4}
\end{figure}

To illustrate the use of the resource, we use it to compare the reported results on two combinations of architecture and data for which there are a significant number of comparisons across different pruning methods, namely: (i) AlexNet and ResNet50 on ImageNet and (ii) ResNet56 and VGG16 on CIFAR10.
Figure~\ref{figure:fig4} show the results reported in terms of the drop in Top-1 accuracy, and percent reduction in FLOPs or parameters where the labels used for the pruning methods are from the primary sources, with suffixes reflecting the variations in pruning methodology used.
 The main observations are that:
\begin{enumerate}
\item For AlexNet on ImageNet,  AOFP-B2 achieves a 41\% reduction in FLOPs with a 0.46\% increase in accuracy and  Dyn Surg \citep{Guo2016}, SSR-L~\citep{Lin2020b} and NeST~\citep{Dai2019} achieve over 93\% reduction in parameters without loss in accuracy. Other methods that compromise accuracy do not necessarily result in a greater reduction in parameters.
\item In comparison to AlexNet, it is harder to prune ResNet50 on ImageNet, although AOFP-C1 achieves 33\% reduction in FLOPs without affecting accuracy. As accuracy is compromised, there are methods that show significant reductions in parameters.  These include KSE and ThinNet with  reductions in parameters by 78\% and 95\%, with a decline of  0.64\%, and 0.84\% in accuracy, respectively.
\item When pruning  ResNet56 on the CIFAR10 data,  the methods KSE and SFP-NFT show reductions in FLOPs of 60\% and 28\% without compromising accuracy. For VGG16 on CIFAR10,  AOFP, PF\_EC  and NetSlimming result in a 75\%, 63\%, and 51\% reduction in flops respectively without reductions in accuracy.  For both networks, it appears to be difficult to gain further reductions (beyond KSE and AOFP) even when compromising accuracy.
\item The charts show that several methods are able to reduce FLOPs and parameters without compromising accuracy and aid generalizability (e.g., AOFP, SFP, NetSlimming), though compromising accuracy a little can sometimes lead to more significant reductions in FLOPs and parameters.  
\item When looking at results within methods, it is possible to confirm our expectation that compromising accuracy can result in greater reductions in parameters and FLOPS (e.g, see results for Filter Sketch and FPGM in Fig.4(b)). However, this tradeoff is not evident when considering results across different pruning methods.
\item Although AOFP does perform well in retaining accuracy for three of the four cases, in general, the performance of the methods varies depending on the architecture and the data set.
\end{enumerate}

\section{CONCLUSION AND FUTURE WORK}

\label{section:conclusion}

This paper has presented a survey of methods for pruning neural networks, focusing on methods based on magnitude pruning, use of  similarity and clustering, and methods based on sensitivity analysis.   

Magnitude based pruning methods have developed from removal of small weights in MLPs to methods for pruning filters and channels which lead to substantial reductions in the size of deep networks. The range of methods developed include: (i) those that  are data dependent and use examples to assess the relevance of output channels, (ii) methods that are independent of data,  which assess the contributions of filters and channels directly, and (iii) methods that utilise optimization algorithms to find filters that approximate channels.  

Methods based on sensitivity analysis are the most transparent in that they are based  on approximating the loss due to changes to a network. The development of methods based on a Taylor series approximation represents the primary line of research in this category of  methods. Different studies have adopted different assumptions in order to make the computation of the Taylor approximation feasible.  In one of the first studies, the OBD method assumed a diagonal Hessian matrix, ignoring both first order gradients and second order non-diagonal gradients. This was followed by OBS, a method that aimed to take  account of non-diagonal elements of the Hessian but has been shown to struggle when pruning multiple weights that are correlated.  The EigenDamage  method aims to take better account of correlations  by approximating the Hessian with a Fisher Information Matrix and using a reparameterization to a new space in which the weights are approximately independent. In an alternative approach, the Collaborative Channel Pruning (CCP) method formulates the pruning task as a quadratic programming problem. \citet{Molchanov2016a} develop a method based on a first-order approximation, arguing that the variance in the loss, as training approaches a local solution, is an indicator of stability and provides a good measure of the importance of filters.  In contrast to most of the other methods that adopt layer by layer pruning with fine tuning after each layer, both EigenDamage and CCP show that it is possible to obtain good results with one-shot pruning followed by fine-tuning. These three recent methods all show good results on large scale networks and data sets, though direct empirical comparisons between them have yet to be published. The survey also found two alternatives to use of Taylor series approximations:  a method that aims to learn which filters to prune \citep{Huang2018a} and a method based on the use of multi-armed bandits \citep{Ameen2020}, both of which have the potential to explore new avenues of research on pruning methods.

The survey reveals a number of positive results about the Lottery Hypothesis:  Lotteries appear to perform well in transfer learning, and lotteries exist for tasks such as NLP and for architectures such as LSTMs. Lotteries even seem to be independent of the type of optimizer used during training. Much of the current research on lotteries is based on deep networks, but it is interesting that one of the earliest papers in the field demonstrates the need to overparameterize a small feedforward network for modelling a 4-bit multiplexor.  Thus, it might  prove fruitful to explore the properties of lotteries on smaller problems as well as the larger networks of today.  The existence of good lotteries does appear to depend on the fine-tuning process adopted and an interesting observation, that challenges some of the empirical studies reported, is that even random pruning can achieve good results following fine-tuning~\citep{Mittal2019}, so further studies of how the remaining weights compensate for those that are removed, could results in new insights. Although studies on lotteries provide valuable insight, further research on specialist hardware and libraries is needed for methods that prune individual weights to become practical~\citep{Elsen2020}.

The survey found least research on methods that use similarity and clustering to develop pruning methods. A method that utilised a cosine similarity measure concluded that it was more suitable for MLPs than CNNs, while a method that utilises agglomerative clustering of filters results in up to a 3-fold reduction on ResNet when it is applied to ImageNet. These results suggest there is merit in developing a more theoretical understanding of the functional equivalence of different classes of deep networks, analogous to the studies on equivalence of MLPs~\citep{kurkova94}.

Given the different approaches to pruning, some may be complimentary, and there is some evidence that combining them might result in further compression of networks.  For example, \citet{He2017}  present results showing that combining their method based on the use of Lasso regression with factorization results in additional gains, and \citet{Han2016b} use a pipeline of magnitude pruning, clustering and Huffman coding to increase the level of compression that can be achieved.  

One of the challenges in making sense of the empirical evaluations reported in the papers surveyed is that, as new deep learning architectures have developed and as new methods have been published, the comparisons carried out have evolved.  The survey has therefore collated the published results of over 50 methods for a variety of data sets and architecture that is available as a resource for other researchers.  Section~\ref{section:resource}
presents a comparison based on this data, highlighting the methods that work well for different architectures on the ImageNet and CIFAR10 data. The comparison of published results suggests that significant reductions can be obtained for AlexNet, ResNet and VGG, though there is no single method that is best, and that it is harder to prune ResNet than the other architectures.  One can hypothesize that this is because its use of skip connections makes it more optimal, though this is something that needs exploring. Likewise, given that different methods seem best for different architectures, it is worth studying and developing methods for specific architectures. The data also reveals that there are limited evaluations on other networks such as the InceptionNet,  DenseNet, SegNet, FCN32 and datasets such as CIFAR100, Flowers102, CUB200-2011 (see Appendix D). A comprehensive independent evaluation of the methods that includes consideration of the issues raised by the Lottery hypothesis across a wider range of data and architectures would be a useful advance in the field.

In conclusion, this survey has presented the key research directions in pruning neural networks by summarizing how the field has progressed from the early algorithms that focused on small fully connected networks to the much larger deep neural networks of today. The survey  has aimed to highlight the motivations and insights identified in the papers, and provides a resource for comparison of the reported results, architectures and data sets used in several studies which we hope will be useful to researchers in the field.\footnote{Resource is available by emailing the first author.}

%\documentclass[smallextended]{svjour3} 
%\documentclass[smallcondensed]{svjour3}     % onecolumn (ditto)
%\documentclass[smallextended]{svjour3}       % onecolumn (second format)
%\documentclass[twocolumn]{svjour3}          % twocolumn
%\usepackage{natbib}
%\usepackage[numbers]{natbib}
% \bibliographystyle{unsrt}
%\usepackage{amsfonts}
%\usepackage{amstext}
%\usepackage{amsmath}
%
%\smartqed  % flush right qed marks, e.g. at end of proof
%
%\usepackage{graphicx}
%
%\usepackage{mathptmx}      % use Times fonts if available on your TeX system
%
% insert here the call for the packages your document requires
%\usepackage{latexsym}
% etc.
%
% please place your own definitions here and don't use \def but
% \newcommand{}{}
%
\newcommand{\rot}{\rotatebox{90}}

%\newcommand{\R}{\mathbb{R}}
% Insert the name of "your journal" with
% \journalname{myjournal}
%
%\begin{document}
%	\title{Methods for Pruning Deep Neural Networks}
%	\author{Sunil Vadera \and Salem Ameen}
%	\institute{S.Vadera, University of Salford, \email{S.Vadera@salford.ac.uk}
%		\and 
%		S.Ameen, University of Salford, \email{S.Ameen@salford.ac.uk}
%	}
%	\date{Received: date / Accepted: date}
%	\maketitle
%\clearpage
\section*{\bf Appendix A: Categorisation of Studies on Pruning}
%\begin{table}[h]
\begin{small}
	\begin{tabular}{|l|l|}\hline \hline
		\multicolumn{2}{l}{\bf Magnitude based pruning methods} \\ \hline
		1988-91& \cite{Kruschke1988, Hanson1989,Weigend1990,Weigend1991} \\ \hline
		2011 & \cite{Graves2011} \\ \hline
		2015 & \cite{Han2015a,Polyak2015}\\ \hline
		2016 &\cite{Guo2016,Hu2016,Wen2016} \\ \hline
		2017 & \cite{Aghasi2017,He2017,Li2017,Liu2017,Luo2017a,Luo2017} \\
		     & \cite{Wang2017,Zhu2017} \\ \hline
		2018 &	\cite{Chen2018,He2018,Huang2018,Lee2018a}\\
		     & \cite{Li2018,Liu2018d,Luo2018,Qin2018}\\
		     & \cite{Yazdani2018,Ye2018, Yu2018, Zhang2018a, Zhou2018} \\ \hline
		2019 &	\cite{Ding2019,Ding2019c,Dettmers2019, Frankle2019,Frankle2019a}   \\ 
			&  \cite{Gui2019, He2019, Helwegen2019, Hou2019, Lee2019, Li2019} \\
			& \cite{Liu2019, Liu2019a, Mittal2019, Morcos2019, Song2019, Xu2019}\\
			& \cite{Yu2019, Zhao2019a,Zhou2019,Zhu2019} \\ \hline
		2020 &	\cite{Hubens2020,Kim2020, Li2020, Lin2020, Lin2020b, Niu2020} \\ \hline
		\multicolumn{2}{l}{\bf Similarity and clustering methods  } \\ \hline
	2017 &	\cite{RoyChowdhury2017}\\ \hline
	2018 &	\cite{Dubey2018, Son2018}\\ \hline
	2019 &	\cite{Ayinde2019, Ding2019b, Li2019b,Li2019sv} \\	\hline 
	2021 & \cite{Lin2021} \\ \hline
	\multicolumn{2}{l}{\bf Sensitivity analysis methods}\\ \hline 
	1988 &	\cite{Mozer1988}\\ \hline
	1990 &	\cite{LeCun1990}\\	\hline
	1992-1993 &	\cite{Hassibi1992,Hassibi1993,Hassibi1993a}\\ \hline
	2006-2009 &  \cite{Xu2006, Endisch2007,Endisch2009} \\ \hline
	2016 &	\cite{Cohen2016,Grosse2016,Molchanov2016a} \\	\hline
	2017 &	 \cite{Ameen2017,Anwar2017, Dong2017}  \\
		 & \cite{Guo2017, Neklyudov2017} \\ \hline
	2018 & \cite{Lin2018,Bao2018,Carreira-Perpinan2018,Jiang2018} \\ 
		 & \cite{Huang2018a,Huynh2018,Zhuang2018} \\ \hline
	2019 &	\cite{Chen2019a,Deng2019,Jin2019,Lee2018} \\
		 &  \cite{Li2019c,Molchanov2019,Peng2019,Qin2019, Wang2019, Xiao2019}	\\ \hline
	2020 & \cite{Ameen2020} \\ \hline
	\multicolumn{2}{l}{\bf Knowledge distillation methods}\\ \hline
	2006  &	\cite{Bucilua2006}  \\	\hline
	2014-2017 &	\cite{Romero2014,Hinton2015,GregorUrban2017} \\	\hline
	2019 &	\cite{Bao2019, Lemaire2019, Zhang2019a, Dong2019}  \\ \hline
	2020 & \cite{kundu2020} \\ \hline
	2021 & \cite{Kaliamoorthi2021} \\ \hline
	\multicolumn{2}{l}{\bf Methods based on rank and reconstruction} \\\hline
	2013-2016 &	\cite{Sainath2013,Jaderberg2014,Lebedev2015,Zhang2016} \\	\hline
	2018 &	\cite{Lin2018a} \\	\hline
	2020 &	\cite{Lin2020c}\\	\hline
	\multicolumn{2}{l}{\bf Quantization methods}\\\hline
	2011 & \cite{Vanhoucke2011}  \\ \hline
	2014-17 &	\cite{Denton2014,Gong2014,Chen2015,Hubara2016,Rastegari2016,Zhou2017a} \\ \hline
	2018 &	\cite{Jacob2018,Krishnamoorthi2018,Liu2018a,Tung2018} \\ \hline
	2019 &  \cite{Banner2019,Chen2019, Jung2019,Zhao2019} \\  \hline
	2020 &  \cite{Wang2020} \\ \hline
	2021 &  \cite{Stock2021} \\ \hline
	\multicolumn{2}{l}{\bf Architectural design methods} \\\hline
	2017 &	\cite{Baker2017, Howard2017, Lin2017, Zoph2017} \\ \hline
	2018 & 	\cite{Gordon2018, He2018a,Hsu2018,liu2018progressive,Pham2018,Zhong2018}\\ \hline
	2019 & \cite{Dai2019,Esteban2019, Liu2019metapruning,Tan2019, Zhang2019b} \\\hline
	2020 & \cite{Elsen2020,Evci2020, Lin2020a} \\	\hline
	\multicolumn{2}{l}{\bf Hybrid Methods}\\ \hline
	2016 &	\cite{Chung2016, Han2016b} \\\hline
	2018-2019 &	\cite{Goetschalckx2018,Gadosey2019} \\ \hline
	\multicolumn{2}{l}{\bf Survey Papers}\\ \hline
	1993 &	\cite{Reed1993} \\ \hline
	2018-2021 &	\cite{Cheng2018a, Cheng2018, Lebedev2018,Elsken2019,Menghani2021} \\ \hline \hline
\end{tabular}
%\caption{Categorisation of studies on pruning neural networks}
\end{small}
%\end{table}

\section*{Appendix B: Summary of Data Sets used in Comparing Pruning Methods}

%\begin{tabular}{|l|l|} \\ \hline
%Data set &	Brief Description  \\ \hline
\begin{description}
	
\item[MNIST~\citep{LeCun1998a}:]
The MNIST (Modified National Institute of Standards and Technology) data set consists of handwritten 28x28 images of digits.  It has 60,000 examples of training data and 10,000 examples for the test set.
\item[PASCAL VOC~\citep{Everingham2015}:] 
The PASCAL VOC data sets have formed the basis of an annual competitions from 2005 to 2012.  The VOC 2007 data annotates objects in 20 classes and consists of 9,963 images and 24,640 annotated objects.  The VOC 2012 data, which consists of 11530 images, are annotated with 27450 regions of interest and 6929 segmentations. 
\item[CamVid~\citep{Brostow2009}:]
CamVid (Cambridge-driving Labelled Video Database) is a data set with videos captured from an automobile. In total over 10mins of video is provided together with over 700 images from the videos that have been labelled.  Each pixel of an image is labelled to indicate if it is part of an objects in one of 32 semantic classes.  
\item[Oxford-Flowers~\citep{Nilsback2008}:]
The Oxford-Flowers data consists of 102 classes of common flowers in the UK.  It provides 2040 training images and 6129 images for testing.
\item[LFW ~\cite{Huang2008}:] The LFW (Labelled Faces in the Wild) is one of the largest and widely used data sets to evaluate face recognition algorithms.  It includes 250x250 pixel images of over 5.7K individuals, with over 13K images in total.
\item[CIFAR-10 \&100 ~\citep{Krizhevsky2009}:]
The CIFAR-10 (Canadian Institute for Advanced Research) data set is a collection of 32x32 colour images in 10 different classes. The data set splits into two sets: 50,000 images for training and 10,000 for testing. CIFAR-100 is similar to CIFAR-10 but has 100 classes, where each class has 500 training images and 100 test images.
\item[ImageNet~\citep{Deng2009}:]    
ImageNet contains millions of images organized using the WordNet hierarchy. It has over 14M images classified in over 21K groups and has provided the data sets for the ImageNet Large Scale Visual Recognition Challenges (ILSVRC) held since 2010.  It is one of the most widely used data sets in benchmarking deep learning models and methods for pruning. A smaller subset known as TinyImageNet is sometimes used and is also available (https://tiny-imagenet.herokuapp.com). It consists of 200 classes with 500 training, 50 validation and 50 testing images per class.  
\item[SVHN~\citep{Netzer2011}:]
The SVHN (Street View House Number) data set is a collection of 600K, 32x32 images of house numbers in Google Street View images. The data set provides 73,257 images for training and 26,032 for testing.
\item[UCSD-Birds~\citep{Wah2011}:]
The UCSD-Birds data set provides 11788 images of birds, labelled as one of 200 different types of species.  The data is split into training and testing sets of 5994 and 5794 respectively.   
\item[Places365~\citep{Zhou2014}:]
Places365 is a data set with 8 million 200x200 pixel images of scenes labeled with one of 434 categories, such as bridge, kitchen, boxing ring, etc.  It provides 50 images per class for validation and the test set consist of 500 images per class. 
\item[CASIA-WebFace~\citep{YiDong2014}:]
Zhen et al. 2014
[170]
CASIA-WebFace is a data set that was created for evaluating face recognition systems.  It provides over 494K  images of over 10K individuals.  
\item[WMT’14 En2De:\citep{Bojar2014}:]
WMT’14 En2De is one of the benchmark language data sets provided for a task set for the Workshop on Statistical Machine Translation held in 2014.   This data set consists of 4.5M English-German pairs of sentences. 
\item[FashionMNIST~\citep{Xiao2017fashionmnist}:]
FashionMNIST is an alternative  to the MNIST data set with  28x28  images of fashion products classified  in 10 categories.  Like MNIST, there are 60,000 images for training and 10,000 images for testing, 
\end{description}

\section*{Appendix C: Summary of Notation}

\begin{itemize}
	
	\item In general, we use $X $ to denote input channels, $W$ to denote weights of filters and $Y$ to denote output channels. 
	\item $Y_{i}$ j  is used to denote the output feature map obtained by applying a filter  $W_{j,i}$  on input channels $X_{i}$ 
	\item $w_{i}$, $w_{j}$, $w_{j,i}$ are used to denote individual weights.
	\item $\beta$ is used to denote a binary mask where $\beta_{i} = 1$ indicates that a feature map or filter should be retained and $\beta=0$ indicates that it should be removed.
	\item $\mathcal{L}$ is used to  denotes a loss function
	\item $L_{0}, L_{1}, L_{2}$ denote norms, with $L_{0}$ counting non-zero values, $L_{1}$, being the sum of absolute. values, $L_{2}$ being the square root of the sum of squares (Euclidean distance). 
	\item $\|W\|_{n}$ will be used to indicate the use of a norm in an equation with the an equation, with the subscript n indicating the specific norm.
	\item $\|W\|_{F}$, known as the Frobenius norm is sometimes used to denote the application of the Euclidean distance to the elements of a matrix
	
\end{itemize}

\section*{Appendix D: Number of Reported Results for a given Architecture and Data Set}
\begin{small}
	
\begin{tabular}{|l|c|c|c|c|c|c|c|c|c|c|c|}   \hline
& \rot{CALTECH256} & \rot{CamVid} & \rot{CIFAR10} & \rot{CIFAR100} &\rot{CUB200-2011} &\rot{Flowers102} & \rot{ImageNet} &\rot{MNIST} &\rot{Pascal VOC} &\rot{SVHN} &\rot{\bf Total} \\ 
		{\bf Architecture} & & & & & & & & & & & \\ \hline
		AlexNet & & & & & & &22 & & & &22 \\ \hline
		DenseNet-100 & & &3 & & & & & & & &3 \\ \hline
		DenseNet-40 & & &3 &1 & & & & & &2 &3 \\ \hline
		FCN32 & & & & & & & & &1 & &1 \\ \hline
		LeNet-300 & & & & & & & &9 & & &9 \\ \hline
		LeNet5 & & & & & & & &23 & & &23 \\ \hline
		Res101 & & & & & & &6 & & & &6 \\ \hline
		Res110 & & &13 & & & & & & & &13 \\ \hline
		Res152 & & & & & & &3 & & & &3 \\ \hline
		Res164 & & &1 &1 & & & & & & &1 \\ \hline
		Res18 &1 & &4 & &1 &1 &3 & & & &8 \\ \hline
		Res20 & & &6 & & & & & & & &6 \\ \hline
		Res32 & & &17 &10 & & & & & & &17 \\ \hline
		Res34 & & & & & & &3 & & & &3 \\ \hline
		Res50 & & & & & & &26 & & & &26 \\ \hline
		Res56 & & &20 & & & &1 & & & &21 \\ \hline
		SegNet  &1 & & & & & & & & &1 &\\ \hline
		VGG16 &1 & &13 &1 &3 &1 &10 & & & &24 \\ \hline
		VGG19 & & &12 &12 &1 & & & & & &13 \\ \hline
		{\bf Total} &1 &1 &48 &13 &3 &1 &59 &28 &1 &2 &115 \\ \hline
\end{tabular}
\end{small}

\clearpage

%\bibliographystyle{num}
%\bibliographystyle{unsrt}
%\bibliography{survey}   % name your BibTeX data base

%\end{document}

  \bibliographystyle{spbasic}      % basic style, author-year citations

  \bibliography{survey}   % name your BibTeX data base

  \end{document}